\documentclass[conference]{IEEEtran}
\IEEEoverridecommandlockouts

\usepackage{cite}
\usepackage{amsmath,amssymb,amsfonts}
\usepackage{algorithmic}
\usepackage{graphicx}
\usepackage{textcomp}
\usepackage{xcolor}
\usepackage{comment}
\usepackage[colorlinks = true, citecolor = black, urlcolor = gray, linkcolor = black]{hyperref}

\usepackage{quiver}
\newcommand\Tau{\mathcal{T}}

\begin{document}

\title{Probing for Consciousness in Machines}

\author{
\IEEEauthorblockN{Mathis Immertreu}
\IEEEauthorblockA{
\textit{CCN Group} \\
\textit{Pattern Recog. Lab.} \\
\textit{FAU Erlangen-Nürnberg}\\
Erlangen, Germany \\
mathis.immertreu@fau.de}
\and
\IEEEauthorblockN{Achim Schilling}
\IEEEauthorblockA{
\textit{Neuroscience Lab,} \\
\textit{University Hospital Erlangen,}\\ 
\textit{CCN Group} \\
\textit{Pattern Recog. Lab.} \\
\textit{FAU Erlangen-Nürnberg}\\
Erlangen, Germany \\
achim.schilling@fau.de\\
}
\and
\IEEEauthorblockN{Andreas Maier}
\IEEEauthorblockA{
\textit{Pattern Recog. Lab.} \\
\textit{FAU Erlangen-Nürnberg}\\
Erlangen, Germany \\
andreas.maier@fau.de}
\and
\IEEEauthorblockN{Patrick Krauss}
\IEEEauthorblockA{
\textit{CCN Group} \\
\textit{Pattern Recog. Lab.} \\
\textit{FAU Erlangen-Nürnberg}\\
Erlangen, Germany \\
patrick.krauss@fau.de}
}


\maketitle

\begin{abstract}
This study explores the potential for artificial agents to develop core consciousness, as proposed by Antonio Damasio's theory of consciousness. According to Damasio, the emergence of core consciousness relies on the integration of a self model, informed by representations of emotions and feelings, and a world model. We hypothesize that an artificial agent, trained via reinforcement learning (RL) in a virtual environment, can develop preliminary forms of these models as a byproduct of its primary task. The agent's main objective is to learn to play a video game and explore the environment. To evaluate the emergence of world and self models, we employ probes—feedforward classifiers that use the activations of the trained agent's neural networks to predict the spatial positions of the agent itself. Our results demonstrate that the agent can form rudimentary world and self models, suggesting a pathway toward developing machine consciousness. This research provides foundational insights into the capabilities of artificial agents in mirroring aspects of human consciousness, with implications for future advancements in artificial intelligence.
\end{abstract}

\begin{IEEEkeywords}
Machine Consciousness, Reinforcement Learning, Core Consciousness, Self Model, World Model, Virtual Environment, Probes
\end{IEEEkeywords}

\section*{Introduction}
As modern computers emerged and their capabilities grew, the idea of machines becoming conscious was already being contemplated, along with the need for methods to investigate this possibility. One of the earliest approaches was the Turing test \cite{10.1093/mind/LIX.236.433}. The basic idea was to let people chat with an unknown partner, which could either be another human or a computer. If humans could no longer distinguish between the two, the machine would be considered conscious. Today, some large language models (LLMs) starting from GPT-4 have arguably passed the Turing test \cite{jones2024peopledistinguishgpt4human}.

However, does this mean that some LLMs have become conscious? John Searle's thought experiment, the Chinese Room \cite{Searle_1980}, addresses this question. In the experiment, a person passes a question written in Chinese into a closed room. Inside, an English-speaking scientist follows a rule book to process Chinese sentences and produce an answer, which is then passed back. Although it appears that the room or the scientist understands Chinese, the scientist is merely following syntactic rules without any comprehension of the meaning. This illustrates that behavior alone is insufficient to conclude that a machine with sophisticated behavior has any understanding or consciousness.

In light of the increasing capabilities and presence of intelligent systems in our daily lives, determining whether machines can become conscious is an increasingly urgent question. Moreover, the exact nature of consciousness remains an unsolved problem, with various contradictory theories about its origins and mechanisms. Notable theories include the controversial integrated information theory \cite{Tononi2016}, which associates consciousness with a maximum of integrated information or cause-effect power, and the global workspace theory \cite{baars2013global}, where parallel information flows compete for access to a global workspace for processing and distribution.

\begin{figure}
     \centering
     \includegraphics[width=.48\textwidth]{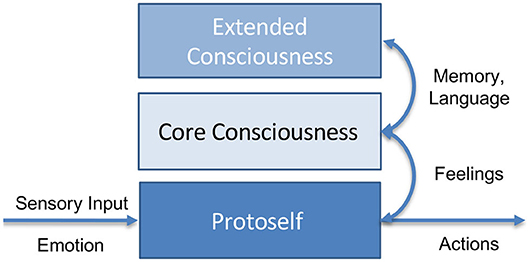}
     \caption{\textbf{Simplified representation of Damasio's model of consciousness (taken from \cite{10.3389/fncom.2020.556544}):} The protoself operates at an unconscious level, processing emotions and sensory input. Core consciousness emerges from the protoself, creating the initial self and world models, allowing the self to relate to its environment. Projections of emotions evolve into higher-order feelings. With access to memory and the integration of complex functions such as language processing, extended consciousness develops, further enhancing the self and world models.}
     \label{fig:damasio}
 \end{figure}

An outlier in these theories is Antonio Damasio's theory of consciousness \cite{damasio2009consciousness}, which provides a detailed mechanism of how consciousness arises, its attributes, and its possible embodiment in the human brain. In \cite{10.3389/fncom.2020.556544} it is argued that this theory is uniquely well-suited for application to artificial intelligence (AI) and machine learning (ML) systems. Damasio structures consciousness into three hierarchical levels: Firstly, the protoself, the neural representation of the body state. Secondly, the core consciousness, a higher-level representation of the self, the world, and their mutual relations, leading to a transient core self, and thirdly, the extended consciousness, which includes memory, language, planning, and other high-level mental activities enabling the continuous autobiographic self.
In this theory, emotions and feelings play a crucial role. Emotions are unconscious reactions to stimuli, and feelings are neural representations of these emotions. Objects can induce emotions, leading to changed feelings and a changed protoself. The combined neural representation of the perceived object and the changes in the protoself due to it forms the core consciousness and creates a core self, a sense of perception belonging to oneself. The autobiographic self, built on top of the core self, includes memories of one's past, consistent characteristics, and future plans. This requires extended consciousness and its functions like memory and planning. From simple reactive emotions to complex plans enabled by extended consciousness, these systems aim to regulate homeostasis, keeping the internal state in a safe range to ensure continued existence and increase the chance of survival. A simplified overview of Damasio's model is illustrated in figure \ref{fig:damasio}.

The concept of emotions, and thus the entire theory, can be applied to AI systems as reactive changes in their embodiment due to stimuli. For robots, this could mean changes in battery levels, actuator positions, or angles, while for smart factories, it might involve changes in cooling systems or production facilities. In computer games, which simulate the world, the agent has a simulated body. Depending on the game, changes in hit points, levels, attributes, resources, or scores can be considered emotions.

This framework allows us to assess whether an AI meets the criteria for different levels of consciousness. \cite{10.3389/fncom.2020.556544} argued that modern algorithms might already be close to achieving "core consciousness." Emotions that move the internal state of the embodiment towards or away from an optimal region can be seen as positive or negative emotions, aligning with the concept of regulating homeostasis. This translates directly to positive and negative rewards in reinforcement learning (RL). RL has demonstrated considerable success in training robots \cite{7989385}, machines \cite{Degrave2022}, and game agents \cite{DBLP:journals/corr/MnihKSGAWR13}, making it a natural choice for training embodied AI systems based on emotions.

Applying Damasio's theory to AI systems offers several advantages over human tests. The entire "brain" activity is accessible, and the whole world and (internal) body state can be easily read and altered, especially in computer games or simulations. Additionally, various experiments can be conducted without significant technical problems or ethical concerns.

We can increasingly narrow down our research questions to testable hypotheses: \\
1. Can a machine become conscious? \\
2. Can a machine possess core consciousness as defined by Damasio? \\
3. Can a machine develop models of the world, itself, and their mutual relations? \\
4. Can an agent in a computer game develop models of the game environment, itself, and their mutual relations? \\

In this context, a world model refers to a neural network that maps external perceptions to an internal representation containing essential structures and dynamics for the agent. A self-model similarly processes perceptions of its internal state. A model sufficient for core consciousness must include both and further model their relationship, such as the boundary between the self and the external world and how changes in the world affect the self and vice versa.

\begin{figure}
    \centering
    \includegraphics[width=.48\textwidth]{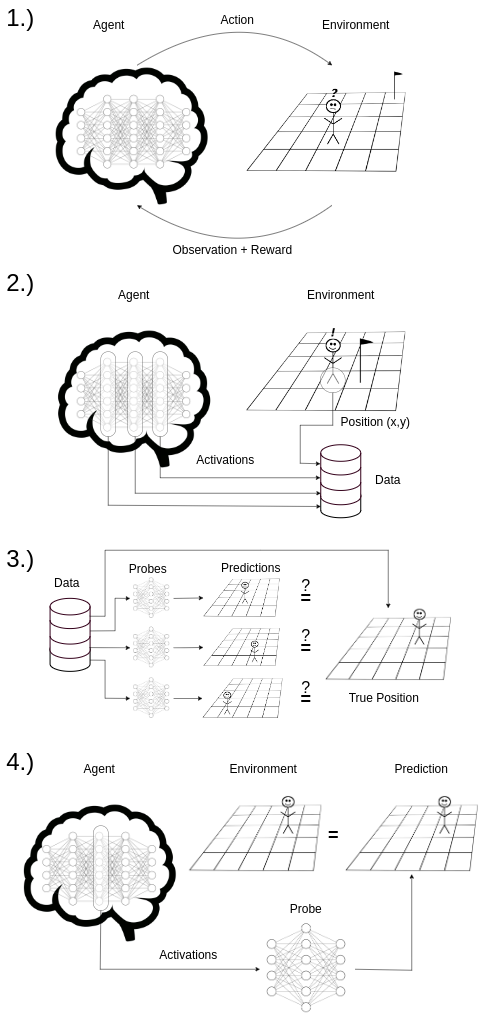}
    \caption{\textbf{Schematized overview of the approach:} 1.) An agent is trained with RL 2.) A dataset of the trained agent's position and neural network's activations is sampled 3.) Using this dataset on each layer's activations a probe is trained to predict the true position. 4.) If one of the probes can predict the true agent position (with an accuracy significantly higher than chance), it shows that the necessary information is contained in the activations. Thus the agent developed a world model.}
    \label{fig:overview}
\end{figure}

In \cite{li2022emergent} it is demonstrated that a transformer trained on sequences of Othello moves developed a world model, effectively modeling the entire board state. Following this idea, we train probes, small classifiers, on the activations of the trained agent to determine if it understands its position in the world. The approach is summarized in Figure \ref{fig:overview}.

\section*{Methods}

\subsection*{Reinforcement Learning}

\begin{figure}
    \centering
\[\begin{tikzcd}
	{\text{Agent}} &&& {\text{Environment}} \\
	& {}
	\arrow["{a_t}", curve={height=-18pt}, from=1-1, to=1-4]
	\arrow["{s_t/o_t}", curve={height=-18pt}, from=1-4, to=1-1]
	\arrow["{r_t}"{description}, from=1-4, to=1-1]
\end{tikzcd}\]
    \caption{\textbf{The basic agent-environment interaction cycle}: The agent observes the current state/observation $s_t/o_t$, decides on an action $a_t$ based on its policy and the environment reacts to this action by returning a reward $r_t$ and the next state/observation $s_{t+1}/o_{t+1}$ beginning the next cycle.}
    \label{fig:RL}
\end{figure}
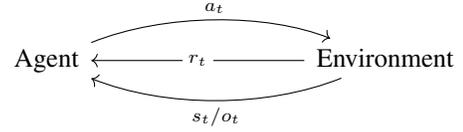

Reinforcement Learning (RL) tries to solve the problem of (optimal) sequential decision making. The basic framework assumes that at every time step $ t \in \mathbb{N}_0$ the agent acts on the environment with an action $a_t\in A$ and the environment returns a state/observation $s_t/o_t \in S/O$ and a reward $r_t\in\mathbb{R}$ to the agent based on transition probabilities $P(S_{t+1}=s' | S_t=s, A_t=a)$ and a reward function $R:S\times A \rightarrow \mathbb{R}$. $A$ denotes the action space, $S$ the state space and $O$ the observation space, which depend on the chosen environment and agent. They can be either continuous or discrete, but for simplicity we focus on the discrete case and assume an episodic setting. The agent learns via this feedback loop to improve its behaviour. The basic RL cycle is illustrated in figure \ref{fig:RL}.

This notion can be further formalized as a (partially observable) Markov decision process. For this we refer to the literature on RL e.g. \cite{Sutton1998}.

The actions of an agent are characterized by a policy which can be either deterministic
\begin{equation*}
    \pi: S \rightarrow A \text{ via } \pi(s)=a
\end{equation*}
 or probabilistic 
 \begin{equation*}
     \pi: S \rightarrow [0,1]^{|A|} \text{ via } \pi(a|s)=P(A=a|S=s).
 \end{equation*}
We denote a policy realized as neural network with weights $\theta$ as $\pi_\theta$.

In this setting the objective is to maximize the expected return, i.e. the expected discounted cumulative reward achieved in an episode,
\begin{equation}\label{obj_fct}
    J(\pi_\theta):=\mathbb{E}_{\pi_\theta}(R(\tau))=\sum_{\tau \in \Tau} P(\tau|\theta) R(\tau)
\end{equation}
with
\begin{equation}
     R(\tau):=\sum_{i=0}^T \gamma^i R(s_t,a_t)
\end{equation}
for trajectories
\begin{equation*}
    \tau=(s_0,a_0,r_0,s_1,a_1,r_1...,s_{T+1}) \in \Tau,
\end{equation*}
the space of trajectories, a discount factor $\gamma \in [0,1]$ and a reward function $R:S\times A \rightarrow \mathbb{R}$ with the probability
\begin{eqnarray}
    &P(\tau|\theta) := \\
    &\rho(s_0) \prod_{t=0}^{T} P(S_{t+1}=s_{t+1} | S_t=s_t, A_t=a_t) \pi_\theta(a_t|s_t),\nonumber
\end{eqnarray}
starting state distribution $\rho$ and episode length $T$.

The discount factor $\gamma<1$ ensures mathematical convergence for $T=\infty$, meaning that the sum of discounted future rewards converges to a finite value even when considering an infinite horizon. This discount factor can be interpreted as modeling uncertainty about the future, as it reduces the impact of rewards that are further away in time, effectively placing more weight on immediate rewards. As a result, the agent prefers closer and more certain rewards compared to those that are further and more uncertain.

\subsection*{Probes}

Probes, a technique from the mechanistic explainability area of AI, are utilized to analyze deep neural networks \cite{alain2018understanding}. They are commonly applied in the field of natural language processing \cite{belinkov-2022-probing}. Probes are typically small, neural network-based classifiers, usually implemented as shallow fully connected networks. They are trained on the activations of specific neurons or layers of a larger neural network to predict certain features, which are generally believed to be necessary or beneficial for the network's task. If probes achieve accuracy higher than chance, it suggests that the information about the feature, or something correlated to it, is present in the activations.

\subsection*{Implementation}

We trained our agents in the NetHack environment using the NetHack Learning Environment (NLE) and MiniHack, a sandbox editor for custom scenarios in NetHack \cite{kuettler2020nethack, samvelyan2021minihack}. NetHack provides a complex, discrete environment with low computational cost. It was first used as a benchmark at the NeurIPS 2021 NetHack challenge, where symbolic methods led by a wide margin \cite{hambro2022insights}. Subsequently, NLE and MiniHack have been used for benchmarking reward modeling with large language model feedback \cite{klissarov2023motif}, automatic curriculum design \cite{parker2022evolving}, internet query usage \cite{nottingham2022learning}, skill transfer \cite{matthews2022skillhack}, and planning with graph-based deep RL \cite{chester2022oracle}. It is also part of a benchmark platform for continual RL \cite{powers2022cora}.

In this paper, we use the MiniHack-Room-Random-15x15-v0 (random), MiniHack-Room-Monster-15x15-v0 (monster), MiniHack-Room-Trap-15x15-v0 (trap), and MiniHack-Room-Ultimate-15x15-v0 (ultimate) maps from MiniHack. The random map consists of a 15x15 grid room with random start (staircase up) and goal (staircase down) positions. The monster map adds 3 monsters, and the trap map adds 15 teleportation traps. The ultimate map includes both features and is unlit, limiting the agent's view to a 3x3 window centered on itself. Teleportation traps are invisible until activated and move the agent to a random free location. All entities and monsters are randomly placed.

The agent can move in all cardinal and ordinal directions and observes the entire map and a 9x9 centered crop as glyphs, unique IDs for every game entity. Later, the action space was restricted to cardinal directions and observations to a 5x5 or 3x3 centered crop. Rewards are given as +1 for reaching the goal and -0.001 per step taken, with a maximum episode length of 300. Further details are in the MiniHack paper \cite{samvelyan2021minihack}.

\begin{figure}
    \centering
    \includegraphics[width=.48\textwidth]{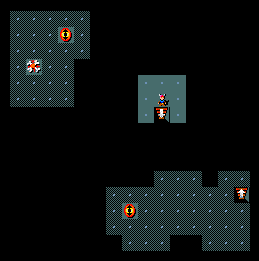}
    \caption{\textbf{An example of the ultimate type map:} The small figure is the agent, the staircases up and down are the start and goal, respectively, the eyes show uncovered teleportation traps and the bones are remains of a defeated monster. The dark grey areas have already been visited by the agent and the light gray 3x3 crop around the agent is the area he just discovered.}
    \label{fig:ultimate}
\end{figure}

\subsection*{Architecture of the agent}

\begin{figure}[ht]
    \centering
    \includegraphics[width=.25\textwidth]{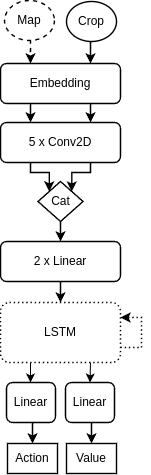}
    \caption{\textbf{A simplified illustration of the agent architecture.}: The map (dashed circle) was only included as input in the first experiment and the LSTM cell (dotted rounded rectangle) became part of the architecture from the second experiment onward.}
    \label{fig:model}
\end{figure}

The basic agent architecture is a simplified version of their baseline model without LSTM cell and the parts to process the message and bottom line status as input e.g. the whole map and a centered crop of the map is given as input, the processing is done by an embedding layer, 5 Conv2D layers and 2 Linear layers followed by two parallel Linear layers with an action (distribution) and the estimated value function of the current state as final output, respectively. The embedding dimension was chosen as 64, each convolution layer contains 16 filters of size 3 besides the last one having only 8 filters. The hidden dimension of the linear layers is 256.
From the second experiment onwards the LSTM cell was added back between the 2 Linear layers and the action and value heads. The cell and hidden state size was chosen as 512.
The architectures are illustrated in figure \ref{fig:model}.

\subsection*{Training}

RLlib \cite{pmlr-v80-liang18b} was used as the RL training framework and the agents were trained with their PPO implementation until convergence. The hyperparameters have been chosen similar to those recommended in the Minihack environment. An agent was trained only on a single map without any restriction of the random seeds. 

\section*{Results}

Subsequent investigations employed probes to infer the spatial coordinates of the agent within the environment, predicated on the neural activations from a single network layer. For this purpose, a comprehensive dataset was compiled, encompassing 230,000 instances per trained agent. These instances were derived by operationalizing the agent within its respective environment, during which both neural activations and corresponding spatial coordinates were documented. The dataset was apportioned into 200,000 samples for training and 30,000 for testing purposes. Training of the probes was oriented towards generating a predictive score for each possible coordinate across a $15 \times 15$ grid (both x and y axes). During the evaluation phase, the predicted coordinates with the highest scores were juxtaposed against the actual positions to assess the predictive accuracy of the probes.

In the initial experiment, a basic agent model without an LSTM cell was used to process inputs from the entire map and a localized $9\times9$ crop. The tests were conducted on two maps: the ultimate map and the trap map, using probes with a single linear layer. The results in Table \ref{tab:results} show accuracies exceeding random chance, especially on the ultimate map, suggesting that the agent's neural activations encode positional information. However, the simplicity of the environment raises questions about whether this is due to direct observations or an internal model. This indicates the need for more complex architectures like RNNs or Transformers. Future experiments should train agents with an LSTM cell using a minimal central crop or use more intricate environments to prevent straightforward observation-based predictions.

In the subsequent experiment, the agent architecture included a $5\times5$ crop as the sole input and an LSTM cell, tested across various maps. The agent's actions were limited to cardinal directions, and map edge positions were excluded to prevent deducing the agent's location from visual input. This setup emphasized the agent's ability to use historical data. Probes, both linear (single layer) and non-linear (three layers with ReLU activation), analyzed the memory aspect by training on the LSTM's hidden and cell states.

The results in Table \ref{tab:results_crop5} show accuracies exceeding chance, more so than in the first experiment. Given the limited $5\times5$ crop input and exclusion of wall observations, positional information must come from the agent's memory and internal representations. Traps and monsters improved accuracies individually but not on the ultimate map, suggesting the need for more quantitative analysis to understand these effects.

The third experiment mirrored the second, except the crop size was reduced to $3\times3$. Probes were trained for 50 epochs using the Adam optimization algorithm, with learning rates set to $0.00005$ for linear probes in the first experiment, and $0.001$ for linear and $0.0001$ for non-linear probes in the second and third experiments. Increased training epochs and advanced learning strategies could potentially enhance performance.

The findings from the third experiment, shown in Table \ref{tab:results_crop3} corroborate prior results, with accuracies significantly above random chance. However, the highest accuracies were on the simplest random map, and the smaller $3\times3$ crop led to lower accuracies on the monster and trap maps. This indicates that a smaller observational input may hinder the agent's ability to accurately infer its position in complex environments.

These results align with preliminary experiments that varied crop sizes, embedding dimensions, hidden/cell state sizes, and convolutional layers across different maps. Accuracies ranged from $17\%$ to $67\%$, consistently above chance levels. The specific agent configuration and the baseline chance level also influenced the final accuracy outcomes.



\begin{table}
    \centering
    \begin{tabular}{c|c|c}
       Layer (output size) \textbackslash Map & Ultimate & Trap\\
       \hline
        1. Conv2D(26544) & 33.9 \% & 8.9 \%\\
        2. Conv2D(26544) & 34.5 \% & 8.8 \%\\
        3. Conv2D(26544) & 34.4 \% & 8.7 \%\\
        4. Conv2D(26544) & 34.2 \% & 8.7 \%\\
        5. Conv2D(13272) & 33.3 \% & 8.9 \%\\
        1. Linear(256) & 29.7 \% & 8.1 \%\\
        2. Linear(256) & 25.3 \% & 7.8 \%
    \end{tabular}
    \caption{\textbf{First experiment's results}: Accuracy of the probes trained on the corresponding activations in predicting the correct x and y values of the position (separately). Chance is $6.7 \%$. The basic agent architecture with map and $9\times9$ crop as input and without LSTM cell was used. The probes consist of one linear layer.}
    \label{tab:results}
\end{table}

\begin{table}
    \centering
    \begin{tabular}{c|c|c|c|c}
       Input (Type) \textbackslash Map & Random & Monster & Trap & Ultimate\\
       \hline
        Hidden state (linear) &  26 \%& 58.8 \% & 40.8 \% & 25.1 \%\\
        Cell state (linear) &  29.9 \%& 62.6 \% & 42.8 \% & 26.4 \%\\
        Hidden state (non-linear) & 31.1  \%& 64.1 \% & 44.4 \% & 27.7 \%\\
        Cell state (non-linear) & 37.3 \% & 67.4 \% & 47.2 \% & 30.5 \%\\
    \end{tabular}
    \caption{\textbf{Second experiment's results}: Accuracy of the probes trained on the hidden and cell states of the LSTM cell in predicting the correct x and y values of the position (separately). Chance is $9.1 \%$. Two rows on the left, right, top and bottom of the map were excluded from the dataset. The agent architecture with only a $5\times5$ crop as input and including the LSTM cell with hidden/cell size 512 was used. The linear probes consist of one linear layer and the non-linear probes of 3 linear layers with ReLUs in between.}
    \label{tab:results_crop5}
\end{table}

\begin{table}
    \centering
    \begin{tabular}{c|c|c|c|c}
       Input (Type) \textbackslash Map & Random & Monster & Trap & Ultimate\\
       \hline
        Hidden state (linear) & 54.8 \% & 49.3 \% & 34 \% & 27.8 \%\\
        Cell state (linear) & 57.4 \% & 50.8 \% & 33.9 \% & 28.8 \%\\
        Hidden state (non-linear) & 58.5 \% & 53.5 \% & 35.4 \% & 29 \%\\
        Cell state (non-linear) & 59.7 \% & 54.8 \% & 36.2 \% & 30.3 \%\\
    \end{tabular}
    \caption{\textbf{Third experiment's results}: Accuracy of the probes trained on the hidden and cell states of the LSTM cell in predicting the correct x and y values of the position (separately). Chance is $7.7 \%$. One row on the left, right, top and bottom of the map was excluded from the dataset. The agent architecture with only a $3\times3$ crop as input and including the LSTM cell with hidden/cell size 512 was used. The linear probes consist of one linear layer and the non-linear probes of 3 linear layers with ReLUs in between.}
    \label{tab:results_crop3}
\end{table}




\section*{Discussion}
Our initial findings suggest that the hidden layer activations encapsulate information regarding the agent's position. Nevertheless, given the simplistic nature of the environment, it remains ambiguous whether this information is directly extracted from observations or assimilated by the agent through a world model. To enhance the efficacy of the method, implementing a more expressive agent architecture, such as RNN or Transformer, coupled with observations that provide less direct information (e.g., a centered crop), is essential.

Consequently, in our subsequent experiments, we trained agents equipped with an LSTM cell, utilizing a narrowly centered crop for observation. Alternatively, introducing a more complex environment could be considered. We also constrained the action space, anticipating that this limitation would foster simpler and more precise latent representations. The findings from our second and third experiments robustly confirm that the agent's position is encoded within the network's activations and indicate that the agent has developed a world model. To rigorously evaluate the impacts and influences of various architectures, environmental settings, and training methodologies, a more detailed and extensive quantitative study is required.

Reducing the crop size complicates the learning challenge but highlights the importance of the agent's ability to infer its position for efficient map navigation. Teleportation traps further obscure the agent's positional accuracy, indicating that lower accuracy might still reflect a more refined world model. A direct comparison of agents could be facilitated by evaluating all agents on a uniform random map, a logical progression for future research. Analysis suggests that the cell state may contain slightly more information than the hidden state, with a non-linear representation enhancing accuracies with non-linear probes. A more comprehensive investigation is needed to thoroughly understand these impacts. However, this paper provides evidence supporting the existence of a world model, with detailed exploration reserved for future studies.

An agent's ability to discern its position may suggest basic core consciousness, but this is not conclusive. Differentiating between a world model and a self-model is crucial. According to Damasio, a self-model is based on stable internal sensations, while a world model relies on variable external observations \cite{damasio2009consciousness}. Future research should train agents with inputs related to physiological states, like hitpoints or experience levels, using changes in these inputs as rewards. For agents with an interpretable world model, modifying specific neural activations, such as those representing the agent's position, can examine their influence on actions, aligning with methods in \cite{li2022emergent}.

The distinction between self-models and world models hinges on stable internal sensations versus variable external observations. However, citation \cite{10.1093/brain/awac194} focuses on differentiating homeostatic feelings and external inputs, which may not directly address internal versus external models.

Future research will have to distinguish between self-models and world models to advance machine consciousness understanding. In addition to an agent consistently appearing in the middle of a crop as a stable element while other fields are chaotic, this might involve incorporating internal state variables into the agent's input and reward structure to observe how these fluctuations affect behavior and decision-making. This will reveal if an agent can develop a true self-model, characterized by stable internal sensations, distinct from variable external observations forming the world model. Analyzing the interplay between these representations is crucial for validating core consciousness per Damasio's framework, laying the groundwork for more sophisticated AI systems.

Future research should use more complex environments to challenge agents' abilities to develop sophisticated self and world models. Current environments are too simplistic. Introducing environments with intricate dynamics, diverse challenges, and multifaceted objectives—such as long-term planning, problem-solving, social interactions, and adaptive learning—will rigorously test agents' potential for consciousness. This will improve evaluation of their ability to form complex internal representations and understand the scalability and generalizability of the models, enhancing the reliability of the findings.

Exploring advanced architectures like transformers and sophisticated RNNs is promising for developing machine consciousness. Current architectures, while effective for initial experiments, may not capture the complexities needed for higher-order cognitive processes. Transformers handle long-range dependencies and parallelize training, while advanced RNNs improve temporal dynamics. These architectures could offer deeper insights into self and world models, enhancing our understanding of how artificial agents develop and utilize complex cognitive functions.

Extending model evaluations to various environments or real-world scenarios is essential for assessing generalizability. While current virtual environments provide controlled settings, real-world applications are more complex. Testing agents in settings like real-time robotics, social simulations, and dynamic ecological systems will offer a comprehensive understanding of their cognitive and adaptive capabilities. This broader evaluation will determine if insights from virtual environments apply to practical scenarios, enhancing relevance and identifying areas for improvement, guiding the development of more robust AI systems.

Our results show that constructing an internal model is crucial for efficiently solving certain tasks. This suggests that even model-free reinforcement learning (RL) approaches might develop implicit internal models, making them not truly model-free. Additionally, some RL exploration strategies require predictions about environmental dynamics, highlighting the practicality of such models. Using this internal model for exploration could further integrate it into the agent's functionality.
In our approach, we used a discount factor for RL to ensure mathematical convergence over an infinite horizon by reducing the impact of distant rewards and emphasizing immediate rewards, effectively modeling future uncertainty. Remarkably, in the context of successor representations (SR) \cite{stachenfeld2017hippocampus, gershman2018successor}, the discount factor plays a similar role by determining how much future states influence the representation of the current state. In particular, in SR the discount factor adjusts the expected discounted future state occupancy, shaping the cognitive map of the environment enabling agents (including humans and animals) to plan and make decisions based on their expectations of future states \cite{stoewer2022neural, stoewer2023neural, stoewer2023conceptual, stoewer2023multi, surendra2023word}. This concept bridges reinforcement learning theories with cognitive science, providing insights into how intelligent behavior emerges from the interaction with the environment. In particular, SR can be seen as a bridge between model-free and model-based approaches \cite{momennejad2017successor, botvinick2019reinforcement}.

This paper underscores the capabilities of AI methodologies, particularly reinforcement learning (RL), in exploring theories of consciousness and advancing explainable AI. Considering AI consciousness is crucial for understanding AI’s intentions and ensuring AI safety. While unconscious AIs can impact outcomes, their lack of awareness means they can't be classified as friendly or evil. Exploring AI consciousness is vital for evaluating the risks and opportunities in AI. Through this work, we aim to contribute to the discourse on whether machines can achieve consciousness.

\begin{thebibliography}{10}
\providecommand{\url}[1]{#1}
\csname url@samestyle\endcsname
\providecommand{\newblock}{\relax}
\providecommand{\bibinfo}[2]{#2}
\providecommand{\BIBentrySTDinterwordspacing}{\spaceskip=0pt\relax}
\providecommand{\BIBentryALTinterwordstretchfactor}{4}
\providecommand{\BIBentryALTinterwordspacing}{\spaceskip=\fontdimen2\font plus
\BIBentryALTinterwordstretchfactor\fontdimen3\font minus \fontdimen4\font\relax}
\providecommand{\BIBforeignlanguage}[2]{{%
\expandafter\ifx\csname l@#1\endcsname\relax
\typeout{** WARNING: IEEEtran.bst: No hyphenation pattern has been}%
\typeout{** loaded for the language `#1'. Using the pattern for}%
\typeout{** the default language instead.}%
\else
\language=\csname l@#1\endcsname
\fi
#2}}
\providecommand{\BIBdecl}{\relax}
\BIBdecl

\bibitem{10.1093/mind/LIX.236.433}
\BIBentryALTinterwordspacing
A.~M. TURING, ``{I.—COMPUTING MACHINERY AND INTELLIGENCE},'' \emph{Mind}, vol. LIX, no. 236, pp. 433--460, 10 1950. [Online]. Available: \url{https://doi.org/10.1093/mind/LIX.236.433}
\BIBentrySTDinterwordspacing

\bibitem{jones2024peopledistinguishgpt4human}
\BIBentryALTinterwordspacing
C.~R. Jones and B.~K. Bergen, ``People cannot distinguish gpt-4 from a human in a turing test,'' 2024. [Online]. Available: \url{https://arxiv.org/abs/2405.08007}
\BIBentrySTDinterwordspacing

\bibitem{Searle_1980}
J.~R. Searle, ``Minds, brains, and programs,'' \emph{Behavioral and Brain Sciences}, vol.~3, no.~3, p. 417–424, 1980.

\bibitem{Tononi2016}
\BIBentryALTinterwordspacing
G.~Tononi, M.~Boly, M.~Massimini, and C.~Koch, ``Integrated information theory: from consciousness to its physical substrate,'' \emph{Nature Reviews Neuroscience}, vol.~17, no.~7, pp. 450--461, Jul 2016. [Online]. Available: \url{https://doi.org/10.1038/nrn.2016.44}
\BIBentrySTDinterwordspacing

\bibitem{baars2013global}
B.~J. Baars, ``A global workspace theory of conscious experience,'' in \emph{Consciousness in philosophy and cognitive neuroscience}.\hskip 1em plus 0.5em minus 0.4em\relax Psychology Press, 2013, pp. 161--184.

\bibitem{damasio2009consciousness}
A.~Damasio and K.~Meyer, ``Consciousness: An overview of the phenomenon and of its possible neural basis,'' \emph{The neurology of consciousness: Cognitive neuroscience and neuropathology}, pp. 3--14, 2009.

\bibitem{10.3389/fncom.2020.556544}
\BIBentryALTinterwordspacing
P.~Krauss and A.~Maier, ``Will we ever have conscious machines?'' \emph{Frontiers in Computational Neuroscience}, vol.~14, 2020. [Online]. Available: \url{https://www.frontiersin.org/articles/10.3389/fncom.2020.556544}
\BIBentrySTDinterwordspacing

\bibitem{7989385}
S.~Gu, E.~Holly, T.~Lillicrap, and S.~Levine, ``Deep reinforcement learning for robotic manipulation with asynchronous off-policy updates,'' in \emph{2017 IEEE International Conference on Robotics and Automation (ICRA)}, 2017, pp. 3389--3396.

\bibitem{Degrave2022}
\BIBentryALTinterwordspacing
J.~Degrave, F.~Felici, J.~Buchli, M.~Neunert, B.~Tracey, F.~Carpanese, T.~Ewalds, R.~Hafner, A.~Abdolmaleki, D.~de~las Casas, C.~Donner, L.~Fritz, C.~Galperti, A.~Huber, J.~Keeling, M.~Tsimpoukelli, J.~Kay, A.~Merle, J.-M. Moret, S.~Noury, F.~Pesamosca, D.~Pfau, O.~Sauter, C.~Sommariva, S.~Coda, B.~Duval, A.~Fasoli, P.~Kohli, K.~Kavukcuoglu, D.~Hassabis, and M.~Riedmiller, ``Magnetic control of tokamak plasmas through deep reinforcement learning,'' \emph{Nature}, vol. 602, no. 7897, pp. 414--419, Feb 2022. [Online]. Available: \url{https://doi.org/10.1038/s41586-021-04301-9}
\BIBentrySTDinterwordspacing

\bibitem{DBLP:journals/corr/MnihKSGAWR13}
\BIBentryALTinterwordspacing
V.~Mnih, K.~Kavukcuoglu, D.~Silver, A.~Graves, I.~Antonoglou, D.~Wierstra, and M.~A. Riedmiller, ``Playing atari with deep reinforcement learning,'' \emph{CoRR}, vol. abs/1312.5602, 2013. [Online]. Available: \url{http://arxiv.org/abs/1312.5602}
\BIBentrySTDinterwordspacing

\bibitem{li2022emergent}
K.~Li, A.~K. Hopkins, D.~Bau, F.~Vi{\'e}gas, H.~Pfister, and M.~Wattenberg, ``Emergent world representations: Exploring a sequence model trained on a synthetic task,'' \emph{arXiv preprint arXiv:2210.13382}, 2022.

\bibitem{Sutton1998}
\BIBentryALTinterwordspacing
R.~S. Sutton and A.~G. Barto, \emph{Reinforcement Learning: An Introduction}, 2nd~ed.\hskip 1em plus 0.5em minus 0.4em\relax The MIT Press, 2018. [Online]. Available: \url{http://incompleteideas.net/book/the-book-2nd.html}
\BIBentrySTDinterwordspacing

\bibitem{alain2018understanding}
G.~Alain and Y.~Bengio, ``Understanding intermediate layers using linear classifier probes,'' 2018.

\bibitem{belinkov-2022-probing}
\BIBentryALTinterwordspacing
Y.~Belinkov, ``Probing classifiers: Promises, shortcomings, and advances,'' \emph{Computational Linguistics}, vol.~48, no.~1, pp. 207--219, Mar. 2022. [Online]. Available: \url{https://aclanthology.org/2022.cl-1.7}
\BIBentrySTDinterwordspacing

\bibitem{kuettler2020nethack}
H.~K{\"{u}}ttler, N.~Nardelli, A.~H. Miller, R.~Raileanu, M.~Selvatici, E.~Grefenstette, and T.~Rockt{\"{a}}schel, ``{The NetHack Learning Environment},'' in \emph{Proceedings of the Conference on Neural Information Processing Systems (NeurIPS)}, 2020.

\bibitem{samvelyan2021minihack}
\BIBentryALTinterwordspacing
M.~Samvelyan, R.~Kirk, V.~Kurin, J.~Parker-Holder, M.~Jiang, E.~Hambro, F.~Petroni, H.~Kuttler, E.~Grefenstette, and T.~Rockt{\"a}schel, ``Minihack the planet: A sandbox for open-ended reinforcement learning research,'' in \emph{Thirty-fifth Conference on Neural Information Processing Systems Datasets and Benchmarks Track (Round 1)}, 2021. [Online]. Available: \url{https://openreview.net/forum?id=skFwlyefkWJ}
\BIBentrySTDinterwordspacing

\bibitem{hambro2022insights}
E.~Hambro, S.~Mohanty, D.~Babaev, M.~Byeon, D.~Chakraborty, E.~Grefenstette, M.~Jiang, J.~Daejin, A.~Kanervisto, J.~Kim \emph{et~al.}, ``Insights from the neurips 2021 nethack challenge,'' in \emph{NeurIPS 2021 Competitions and Demonstrations Track}.\hskip 1em plus 0.5em minus 0.4em\relax PMLR, 2022, pp. 41--52.

\bibitem{klissarov2023motif}
M.~Klissarov, P.~D'Oro, S.~Sodhani, R.~Raileanu, P.-L. Bacon, P.~Vincent, A.~Zhang, and M.~Henaff, ``Motif: Intrinsic motivation from artificial intelligence feedback,'' \emph{arXiv preprint arXiv:2310.00166}, 2023.

\bibitem{parker2022evolving}
J.~Parker-Holder, M.~Jiang, M.~Dennis, M.~Samvelyan, J.~Foerster, E.~Grefenstette, and T.~Rockt{\"a}schel, ``Evolving curricula with regret-based environment design,'' in \emph{International Conference on Machine Learning}.\hskip 1em plus 0.5em minus 0.4em\relax PMLR, 2022, pp. 17\,473--17\,498.

\bibitem{nottingham2022learning}
K.~Nottingham, A.~Pyla, S.~Singh, and R.~Fox, ``Learning to query internet text for informing reinforcement learning agents,'' \emph{arXiv preprint arXiv:2205.13079}, 2022.

\bibitem{matthews2022skillhack}
M.~Matthews, M.~Samvelyan, J.~Parker-Holder, E.~Grefenstette, and T.~Rockt{\"a}schel, ``Skillhack: A benchmark for skill transfer in open-ended reinforcement learning,'' in \emph{ICLR Workshop on Agent Learning in Open-Endedness}, 2022.

\bibitem{chester2022oracle}
A.~Chester, M.~Dann, F.~Zambetta, and J.~Thangarajah, ``Oracle-sage: Planning ahead in graph-based deep reinforcement learning,'' in \emph{Joint European Conference on Machine Learning and Knowledge Discovery in Databases}.\hskip 1em plus 0.5em minus 0.4em\relax Springer, 2022, pp. 52--67.

\bibitem{powers2022cora}
S.~Powers, E.~Xing, E.~Kolve, R.~Mottaghi, and A.~Gupta, ``Cora: Benchmarks, baselines, and metrics as a platform for continual reinforcement learning agents,'' in \emph{Conference on Lifelong Learning Agents}.\hskip 1em plus 0.5em minus 0.4em\relax PMLR, 2022, pp. 705--743.

\bibitem{pmlr-v80-liang18b}
\BIBentryALTinterwordspacing
E.~Liang, R.~Liaw, R.~Nishihara, P.~Moritz, R.~Fox, K.~Goldberg, J.~Gonzalez, M.~Jordan, and I.~Stoica, ``{RL}lib: Abstractions for distributed reinforcement learning,'' in \emph{Proceedings of the 35th International Conference on Machine Learning}, ser. Proceedings of Machine Learning Research, J.~Dy and A.~Krause, Eds., vol.~80.\hskip 1em plus 0.5em minus 0.4em\relax PMLR, 10--15 Jul 2018, pp. 3053--3062. [Online]. Available: \url{https://proceedings.mlr.press/v80/liang18b.html}
\BIBentrySTDinterwordspacing

\bibitem{10.1093/brain/awac194}
\BIBentryALTinterwordspacing
A.~Damasio and H.~Damasio, ``{Homeostatic feelings and the biology of consciousness},'' \emph{Brain}, vol. 145, no.~7, pp. 2231--2235, 05 2022. [Online]. Available: \url{https://doi.org/10.1093/brain/awac194}
\BIBentrySTDinterwordspacing

\bibitem{stachenfeld2017hippocampus}
K.~L. Stachenfeld, M.~M. Botvinick, and S.~J. Gershman, ``The hippocampus as a predictive map,'' \emph{Nature neuroscience}, vol.~20, no.~11, pp. 1643--1653, 2017.

\bibitem{gershman2018successor}
S.~J. Gershman, ``The successor representation: its computational logic and neural substrates,'' \emph{Journal of Neuroscience}, vol.~38, no.~33, pp. 7193--7200, 2018.

\bibitem{stoewer2022neural}
P.~Stoewer, C.~Schlieker, A.~Schilling, C.~Metzner, A.~Maier, and P.~Krauss, ``Neural network based successor representations to form cognitive maps of space and language,'' \emph{Scientific Reports}, vol.~12, no.~1, p. 11233, 2022.

\bibitem{stoewer2023neural}
P.~Stoewer, A.~Schilling, A.~Maier, and P.~Krauss, ``Neural network based formation of cognitive maps of semantic spaces and the putative emergence of abstract concepts,'' \emph{Scientific Reports}, vol.~13, no.~1, p. 3644, 2023.

\bibitem{stoewer2023conceptual}
------, ``Conceptual cognitive maps formation with neural successor networks and word embeddings,'' in \emph{2023 IEEE International Conference on Development and Learning (ICDL)}.\hskip 1em plus 0.5em minus 0.4em\relax IEEE, 2023, pp. 391--395.

\bibitem{stoewer2023multi}
------, ``Multi-modal cognitive maps based on neural networks trained on successor representations,'' \emph{arXiv preprint arXiv:2401.01364}, 2023.

\bibitem{surendra2023word}
K.~Surendra, A.~Schilling, P.~Stoewer, A.~Maier, and P.~Krauss, ``Word class representations spontaneously emerge in a deep neural network trained on next word prediction,'' in \emph{2023 International Conference on Machine Learning and Applications (ICMLA)}.\hskip 1em plus 0.5em minus 0.4em\relax IEEE, 2023, pp. 1481--1486.

\bibitem{momennejad2017successor}
I.~Momennejad, E.~M. Russek, J.~H. Cheong, M.~M. Botvinick, N.~D. Daw, and S.~J. Gershman, ``The successor representation in human reinforcement learning,'' \emph{Nature human behaviour}, vol.~1, no.~9, pp. 680--692, 2017.

\bibitem{botvinick2019reinforcement}
M.~Botvinick, S.~Ritter, J.~X. Wang, Z.~Kurth-Nelson, C.~Blundell, and D.~Hassabis, ``Reinforcement learning, fast and slow,'' \emph{Trends in cognitive sciences}, vol.~23, no.~5, pp. 408--422, 2019.

\end{thebibliography}
\end{document}